%% file: main.tex
\documentclass[10pt,twocolumn,letterpaper]{article}

\usepackage{multirow}
\usepackage[pagenumbers]{cvpr} 

\input{preamble}
\definecolor{cvprblue}{rgb}{0.21,0.49,0.74}
\usepackage[pagebackref,breaklinks,colorlinks,allcolors=cvprblue]{hyperref}

\def\paperID{*****} 
\def\confName{CVPR}
\def\confYear{2026}

\title{Beyond the Mean: Modelling Annotation Distributions in Continuous Affect Prediction}

\author{Kosmas Pinitas\\
Department of Digital Systems\\
University of Piraeus\\
{\tt\small kpinitas@unipi.gr}
\and
Ilias Maglogiannis\\
Department of Digital Systems\\
University of Piraeus\\
{\tt\small imaglo@unipi.gr}
}

\begin{document}
\maketitle
\input{sec/0_abstract}    
\input{sec/1_intro}
\input{sec/rel_work}
\input{sec/methodology}
\input{sec/experiments}

{
    \small
    \bibliographystyle{ieeenat_fullname}
    \bibliography{main}
}


\end{document}

%% file: sec/0_abstract.tex
\begin{abstract}
Emotion annotation is inherently subjective and cognitively demanding, producing signals that reflect diverse perceptions across annotators rather than a single ground truth. In continuous affect prediction, this variability is typically collapsed into point estimates such as the mean or median, discarding valuable information about annotator disagreement and uncertainty. In this work, we propose a distribution-aware framework that models annotation consensus using the Beta distribution. Instead of predicting a single affect value, models estimate the mean and standard deviation of the annotation distribution, which are transformed into valid Beta parameters through moment matching. This formulation enables the recovery of higher-order distributional descriptors, including skewness, kurtosis, and quantiles, in closed form. As a result, the model captures not only the central tendency of emotional perception but also variability, asymmetry, and uncertainty in annotator responses. We evaluate the proposed approach on the SEWA and RECOLA datasets using multimodal features. Experimental results show that Beta-based modelling produces predictive distributions that closely match the empirical annotator distributions while achieving competitive performance with conventional regression approaches. These findings highlight the importance of modelling annotation uncertainty in affective computing and demonstrate the potential of distribution-aware learning for subjective signal analysis.

\end{abstract}

%% file: sec/1_intro.tex
\section{Introduction}
\label{sec:intro}

Human emotion is one of the most complex signals to model. It is inherently multimodal, expressed through speech, facial expressions,  physiological responses, and its perception requires substantial cognitive effort from human annotators \cite{fatigue}. Unlike deterministic signals with a well-defined ground truth, emotional perception is inherently subjective. Different observers may interpret the same behavioural cues differently depending on context, cultural background, and perceptual bias. As a result, emotion annotations rarely converge to a single value; instead, they form distributions that reflect diverse interpretations of affective behaviour \cite{picard2000affective}.

In affective computing, this variability has traditionally been treated as noise around a latent true label. However, recent perspectives suggest that annotator disagreement itself carries meaningful information about emotional perception. Multiple annotators observing the same stimulus may assign different affective values, reflecting ambiguity or competing interpretations of the observed behaviour. From this perspective, disagreement is not merely an artefact of annotation but an intrinsic property of emotional signals. Therefore, modelling this variability provides an opportunity to capture richer information about how emotions are perceived by humans.

Despite this, most continuous emotion recognition pipelines reduce annotations to point estimates such as the mean or median \cite{pinitas2024varying,nicolaou2011continuous,pinitas2025privileged}. While this simplification facilitates standard regression frameworks, it collapses the underlying annotation distribution into a single scalar target. In doing so, important characteristics of the annotation signal, such as variability, asymmetry, and uncertainty, are lost. Although widely adopted, such averaging implicitly assumes that disagreement reflects noise rather than meaningful variation. Previous studies have demonstrated that overlooking annotator variability can impair both generalization and calibration in affective models \cite{cowie2001emotion, schuller2009acoustic, pinitas2024varying}. This issue is particularly relevant for real-world applications in healthcare and education, as exemplified by European initiatives like AI4Work \cite{maglogiannis2024ai4work}, which employs Digital Twin technology to predict burnout symptoms and detect stress in demanding work environments. 

These limitations motivate modelling frameworks that explicitly represent the distributional nature of emotion annotations. Rather than predicting a single consensus value, distribution-aware approaches aim to capture both central tendency and variability in human perception. Such representations can provide richer insight into affective behaviour, allowing models to distinguish between confident emotional signals and inherently ambiguous ones. In domains involving subjective human judgements, modelling the full annotation distribution may therefore lead to more robust and interpretable systems.

In this paper, we propose a distribution-centric framework for affective consensus modelling based on the Beta distribution. Our contributions are threefold: (i) we introduce a Beta-based formulation in which models predict mean and standard deviation to obtain valid Beta parameters; (ii) we show how this formulation enables closed-form computation of higher-order descriptors such as skewness, kurtosis, and quantiles; (iii) we demonstrate through experiments on SEWA \cite{kossaifi2019sewa} and RECOLA \cite{ringeval2013recola} that Beta-based models match or surpass point-estimate baselines while capturing richer distributional properties of annotator disagreement. This shows that reliable higher-order statistics can be derived from simple $\mu,\sigma$ predictions, without the need to regress each descriptor directly. Our findings establish emotion as a benchmark for probabilistic signal processing and show the broader applicability of Beta-based consensus modelling in subjective domains.

%% file: sec/rel_work.tex
\section{Related Work}

Emotion annotations are inherently subjective and often exhibit disagreement across annotators. 
Rather than treating this variability as noise, several approaches have attempted to explicitly 
model annotation ambiguity in affective computing. Early work explored interval-based methods 
such as strength modelling \cite{han2017strength}, which represents annotator agreement through 
confidence intervals. Other approaches incorporate uncertainty directly into regression 
frameworks. For instance, Gaussian-based formulations have been used to estimate both mean and 
variance of affect signals, sometimes incorporating temporal lag correction or additional 
constraints \cite{kollias2021affect, yang2024eeg}. 

More recently, probabilistic learning frameworks have been proposed to capture uncertainty in 
emotion prediction. Bayesian neural networks model predictive distributions over outputs 
\cite{prabhu2021end}, while evidential regression approaches estimate distribution parameters 
directly from data \cite{amini2020deep, wu2023estimating}. Mixture density networks have also 
been explored to represent multi-modal output distributions \cite{dang2017investigation}. 
These methods allow models to capture uncertainty in the prediction process, but they are often 
designed around Gaussian assumptions or focus primarily on modelling the mean and variance of 
annotations.

Similar trends appear in related perceptual modelling tasks. In speech quality estimation, 
MBNet models listener bias during mean opinion score (MOS) prediction \cite{leng2021mbnet}, 
while DeePMOS predicts the full MOS distribution rather than a single score 
\cite{liang2023deepmos}. Variance-aware regression approaches further attempt to recover 
uncertainty in subjective labels \cite{faridee2022predicting}. Beyond Gaussian assumptions, 
recent work has explored alternative distributional formulations such as Beta-based quantile 
regression \cite{akrami2024beta} and conformalised regression for uncertainty estimation 
\cite{parente2024conformalized}. These studies highlight the benefits of treating subjective 
labels as distributions rather than single-valued targets.

Despite this progress, most existing approaches remain limited to modelling the mean and 
variance of annotation distributions. Higher-order characteristics of subjective perception, 
such as skewness, kurtosis, entropy, or distribution quantiles, are rarely considered. In this 
work, we address this limitation through a Beta-based modelling framework that predicts the 
mean and variance of annotation signals and derives higher-order descriptors in closed form. 
This formulation is lightweight and model-agnostic, and complements recent advances in ordinal 
emotion modelling \cite{wu2025ordinal} and Beta-based regression \cite{akrami2024beta}. 
Importantly, it allows us to explicitly examine whether predicted distributions capture the 
structure of annotator disagreement rather than collapsing it into a single consensus value.

\begin{figure*}[!tb]
    \centering
    \includegraphics[width=0.9\linewidth]{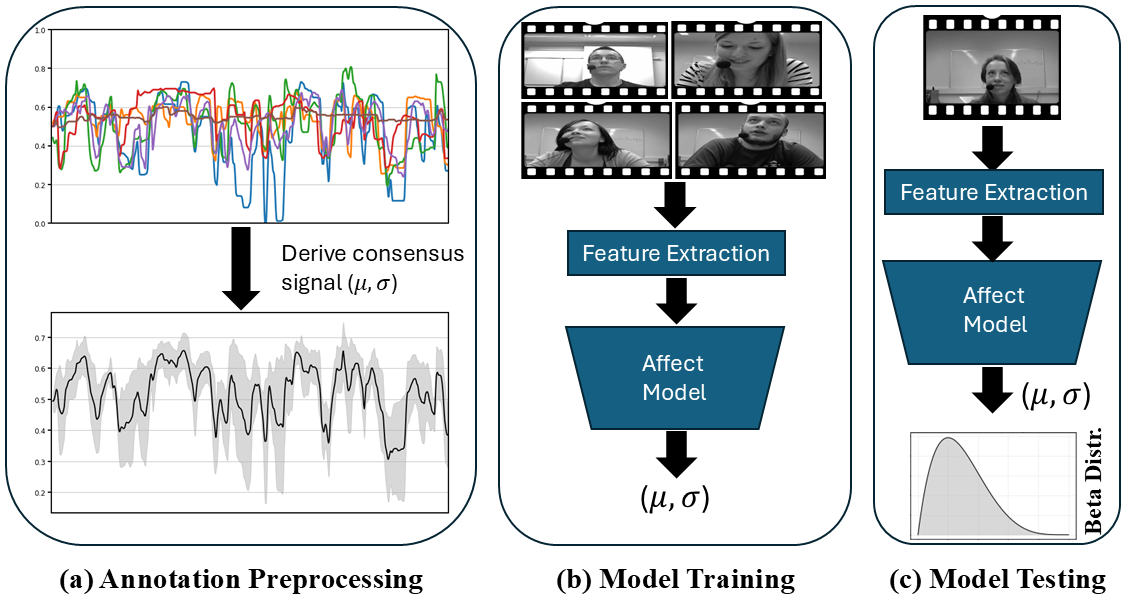}
  \caption{\textbf{Framework Overview:} 
(a) annotator signals form a probabilistic consensus $(\mu,\sigma)$; 
(b) multimodal features are mapped to these parameters during training; 
(c) at test time, predicted $(\mu,\sigma)$ define a Beta distribution capturing annotation ambiguity.}

    \label{fig:concept_figure}
\end{figure*}

%% file: sec/methodology.tex
\section{Methodology}

\subsection{Beta Distribution for Annotation Consensus Modelling}

Continuous affect annotations are inherently ambiguous, as different annotators may perceive the same emotional behaviour differently. Rather than collapsing these perceptions into a single consensus value, we model the distribution of annotator responses directly. Specifically, we assume that the distribution of annotations for a given time window can be approximated by a Beta distribution.

The Beta distribution is defined on the bounded interval $[0,1]$, making it well suited for modelling normalised affect dimensions such as valence and arousal. In our experiments, annotation signals are rescaled to this interval prior to modelling. The flexibility of the Beta distribution allows it to represent a wide range of shapes, including symmetric, skewed, and concentrated distributions, which makes it suitable for capturing the variability and asymmetry often observed in subjective annotations. Importantly, its parameters $(\alpha,\beta)$ allow closed-form computation of descriptive statistics such as mean, variance, skewness, kurtosis, and quantiles.

Let $\{a_i\}_{i=1}^N$ denote annotations from $N$ raters at a given time step. The empirical mean $\mu$ and variance $\sigma^2$ of these annotations are used to recover the parameters of the Beta distribution through moment matching:
\begin{equation}
    \phi = \frac{\mu(1-\mu)}{\sigma^2} - 1, \qquad
    \alpha = \mu \phi, \qquad
    \beta = (1-\mu)\phi.
\end{equation}

This mapping is valid under the conditions
\begin{equation}
    0 < \mu < 1, \qquad
    0 < \sigma^2 < \mu(1-\mu), \qquad
    \alpha > 0, \qquad
    \beta > 0,
\end{equation}
which ensure that $(\alpha,\beta)$ define a valid Beta distribution. In practice, small numerical adjustments can be applied when necessary to guarantee valid parameter values.

Once the Beta parameters are obtained, the resulting distribution provides a compact representation of annotator consensus and disagreement. In particular, higher-order descriptors of the annotation distribution can be computed analytically. The skewness captures asymmetry in annotator responses, while the excess kurtosis describes the concentration of annotations around the mean:

\[
\mathrm{Skew}(X) =
\frac{2(\beta-\alpha)\sqrt{\alpha+\beta+1}}
{(\alpha+\beta+2)\sqrt{\alpha\beta}},
\]

\[
\mathrm{Kurt}_{\mathrm{ex}}(X) =
\frac{6\!\left[(\alpha-\beta)^2(\alpha+\beta+1)-\alpha\beta(\alpha+\beta+2)\right]}
{\alpha\beta(\alpha+\beta+2)(\alpha+\beta+3)}.
\]

Finally, distribution quantiles—including the median—can be obtained by inverting the regularised incomplete Beta function:

\[
q_p = I^{-1}_p(\alpha,\beta).
\]

For $\alpha,\beta > 1$, a useful closed-form approximation for the median is

\[
(\alpha-1/3)/(\alpha+\beta-2/3).
\]

This formulation allows models to predict only the mean and variance of the annotation distribution while recovering richer descriptors of annotator perception in closed form.

\subsection{Machine Learning Models}

This study employs lightweight two-layer ANNs with ReLU activations, where each layer reduces dimensionality by 25\% (first layer at 75\% of input size, second at 50\%). Models are trained to predict $\mu$ and $\sigma$ from multimodal features under three variants: $M_I$, two independent networks for $\mu$ and $\sigma$; $M_S$, a shared first layer with split second layers; and $M_F$, a fully shared network with two outputs. These designs assess the trade-off between independent modelling and parameter sharing. As a baseline ($B$), we implement point-regressors that train a separate network for each descriptor (e.g., $\mu$, $\sigma$, skewness, kurtosis, quantiles), directly predicting values without probabilistic coupling. In contrast, our Beta-based models require predicting only $\mu$ and $\sigma$, from which all higher-order descriptors are obtained in closed form, offering a more efficient and coherent approach to consensus modelling.

\section{Datasets and Preprocessing}

\subsection{Datasets}

We evaluate the proposed framework on two widely used benchmarks for continuous affect prediction: RECOLA \cite{ringeval2013recola} and SEWA \cite{kossaifi2019sewa}. Both datasets provide multimodal recordings together with continuous annotations of emotional dimensions in the valence–arousal space, making them suitable benchmarks for studying annotation variability in affective computing.

\textbf{RECOLA} (Remote Collaborative and Affective interactions) contains multimodal recordings of dyadic interactions between participants collaborating on a remote task. The dataset includes synchronised audio, video, and physiological signals collected from 23 participants during a collaborative problem-solving activity. Continuous annotations of valence and arousal were obtained from six expert raters who independently evaluated the recordings while watching the interaction videos. The dataset provides several feature modalities including 40 acoustic descriptors, 130 visual features derived from facial behaviour analysis, and 116 physiological features extracted from biosignals such as electrodermal activity and electrocardiography. RECOLA has become a standard benchmark for continuous emotion recognition due to its rich multimodal recordings and carefully curated annotations.

\textbf{SEWA} (Sentiment Analysis in the Wild) is a large-scale audiovisual dataset designed to capture spontaneous emotional behaviour in naturalistic settings. The dataset contains more than 2,000 minutes of audio–visual recordings from 398 participants spanning six different cultures (British, German, Hungarian, Greek, Serbian, and Chinese) with ages ranging from 18 to 65 years. Participants were recorded while watching advertisements and subsequently discussing them through video-chat interactions. The dataset includes continuous annotations of several affective dimensions including valence, arousal, liking, and agreement. For the purposes of this work, we focus on valence and arousal annotations provided by three annotators. In addition to the raw recordings, the dataset provides a rich set of behavioural features including facial landmarks, facial action units, and vocalisation cues extracted from the audiovisual streams. 

\subsection{Preprocessing}

For both datasets, recordings were segmented into temporal windows of 3 seconds with a shift of 400\,ms in order to generate temporally consistent samples for model training \cite{pinitas2022supervised, makantasis2023lab}. This sliding-window strategy is commonly used in continuous affect prediction as it balances temporal resolution with annotation stability. The chosen window size allows the model to capture short-term behavioural patterns while smoothing very rapid fluctuations in the signals. The 400\,ms stride produces overlapping windows, which increases the number of training samples and preserves temporal continuity in the data.

Within each window, multimodal features were aggregated by computing their average value over the corresponding time interval. This aggregation step produces a single feature vector per window while preserving the overall behavioural trends within the segment. Since the feature extraction pipelines of RECOLA and SEWA provide frame-level descriptors, this averaging operation also ensures temporal alignment across modalities by mapping all signals to a common time scale.

Annotation traces were processed using the same windowing scheme to ensure consistency between input features and supervisory signals. For each time window, the annotation values provided by all raters were first temporally aligned and aggregated. We then computed the empirical mean and standard deviation across annotators. The mean reflects the central tendency of perceived emotion, while the standard deviation captures the level of disagreement among annotators. This representation therefore provides a compact description of both consensus and ambiguity in emotional perception. Each time window is thus associated with a pair $(\mu,\sigma)$ describing the distribution of annotator responses.

Since the proposed framework models annotation distributions using the Beta distribution, which is defined on the interval $[0,1]$, all annotation signals were linearly rescaled to this range prior to modelling. This normalisation step preserves the relative relationships between annotations while ensuring compatibility with the Beta parameterisation. After applying the preprocessing pipeline, the resulting datasets contain 5268 samples for RECOLA and 8162 samples for SEWA.

%% file: sec/experiments.tex
\section{Experiments}

\subsection{Training and Evaluation Protocol}

In this initial study, we evaluate the proposed models on continuous arousal and valence regression using five-fold subject-independent cross-validation. This protocol ensures that training, validation, and test sets contain recordings from non-overlapping participants, providing a realistic estimate of model generalisation to unseen subjects (see Figure~\ref{fig:concept_figure}). Neural networks are trained using the Adam optimiser with a learning rate of $10^{-3}$ and a batch size of 128 for up to 50 epochs. Early stopping is applied based on the validation loss using mean squared error (MSE) as the optimisation objective, with training terminated after five epochs without improvement.  To account for variability due to random initialisation, each experiment is repeated 10 times using different random seeds. Model performance is evaluated using the Concordance Correlation Coefficient (CCC), which measures agreement between predicted and ground-truth signals by jointly accounting for correlation and scale differences. Statistical significance is assessed using paired Wilcoxon signed-rank tests with a significance level of $p < 0.05$.

\subsection{Results}

\begin{table}[tbh!]
\centering
\caption{CCC performance on RECOLA and SEWA for arousal and valence regression. 
$M_I$: independent $\mu,\sigma$ networks; 
$M_S$: shared first layer with split second layers; 
$M_F$: fully shared network with two outputs. 
Best scores in bold; non-significant differences also marked.}
\centering
\resizebox{0.67\columnwidth}{!}{%
\begin{tabular}{l||@{ }c@{ }|@{ }c@{ }|@{ }c@{ }|@{ }c@{ }|@{ }c@{ }}
%

\multicolumn{2}{c}{\textbf{RECOLA}} & 
\multicolumn{2}{c}{\textbf{Arousal}} &
\multicolumn{2}{c}{\textbf{Valence}} \\ 
\hline \hline
Modality& Model& $\mu$ & $\sigma$ 
& $\mu$ & $\sigma$  
\\ \hline
\multirow{3}{*}{Audio} 
& $M_I$ & \textbf{0.19} & \textbf{0.04} & \textbf{0.54} & \underline{0.25}  \\ \cline{2-6}
& $M_S$ & \underline{0.17} & \underline{0.04} & \underline{0.51} & \textbf{0.27}\\  \cline{2-6}
& $M_F$ & 0.13 & 0.02 & \underline{0.52} & 0.22\\  \hline\hline
\multirow{3}{*}{Visual} 
& $M_I$ & \textbf{0.05} & \textbf{0.03} & \textbf{0.03} & \textbf{0.03}  \\ \cline{2-6}
& $M_S$ & \underline{0.03} & \underline{0.01} & \underline{0.02} & \underline{0.03}\\  \cline{2-6}
& $M_F$ & \underline{0.03} & \underline{0.01} & \underline{0.03} & \underline{0.02}\\  \hline\hline
\multirow{3}{*}{Physiology}
& $M_I$ & 0.11 & \underline{0.01} & 0.14 & \underline{0.14}   \\ \cline{2-6}
& $M_S$ & \textbf{0.15} & \textbf{0.01} & \underline{0.18} & \textbf{0.14} \\  \cline{2-6}
& $M_F$ & \underline{0.13} & \underline{0.01} & \textbf{0.20} & \underline{0.12} \\  \hline\hline
\multirow{3}{*}{Fusion}
& $M_I$ & \textbf{0.24} & 0.01 & \underline{0.48} & \underline{0.26}  \\ \cline{2-6}
& $M_S$ & 0.21 & \underline{0.05} & 0.47 & \textbf{0.28}\\  \cline{2-6}
& $M_F$ & \underline{0.23} & \textbf{0.07} & \textbf{0.50} & \underline{0.26}\\  \hline\hline

\multicolumn{2}{c}{\textbf{SEWA}} & 
\multicolumn{2}{c}{\textbf{Arousal}} &
\multicolumn{2}{c}{\textbf{Valence}} \\ 
\hline \hline
Modality& Model& $\mu$ & $\sigma$ 
& $\mu$ & $\sigma$  
\\ \hline
\multirow{3}{*}{Audio} 

& $M_I$ & \textbf{0.02} & \textbf{0.03} & \textbf{0.04} & \textbf{0.01}  \\ \cline{2-6}
& $M_S$ & \underline{0.01} & \underline{0.01} & \underline{0.01} & \underline{0.01}\\  \cline{2-6}
& $M_F$ & \underline{0.01} & \underline{0.01} & \underline{0.02} & \underline{0.01}\\  \hline\hline
\multirow{3}{*}{Visual} 
& $M_I$ & 0.77 & 0.55 & \textbf{0.76} & \textbf{0.53}  \\ \cline{2-6}
& $M_S$ & 0.75 & \underline{0.60} & \underline{0.75} & \underline{0.51}\\  \cline{2-6}
& $M_F$ & \textbf{0.80} & \textbf{0.61} & \underline{0.76} & \underline{0.51}\\  \hline\hline

\multirow{3}{*}{Fusion}
& $M_I$ & \underline{0.76} & \underline{0.61} & 0.74 & 0.50  \\ \cline{2-6}
& $M_S$ & \textbf{0.78} & \textbf{0.65} & 0.73 & 0.52\\  \cline{2-6}
& $M_F$ & \underline{0.76} & \underline{0.65} & \textbf{0.78} & \textbf{0.57}\\  \hline\hline

\end{tabular}
}
\label{table:models}
\end{table}

\begin{table}[tbh!]
\centering
\caption{KL divergence of predicted distributions against ground-truth annotation Betas on RECOLA and SEWA. 
$\mathcal{U}$: Uniform baseline; 
$\mathcal{B}$: true Beta distribution. 
Lower values indicate closer match to annotator distributions.}
\centering
\resizebox{0.65\columnwidth}{!}{%
\begin{tabular}{l||@{ }c@{ }|@{ }c@{ }|@{ }c@{ }|@{ }c@{ }|@{ }c@{ }}
%

\multicolumn{2}{c}{\textbf{RECOLA}} & 
\multicolumn{2}{c}{\textbf{Arousal}} &
\multicolumn{2}{c}{\textbf{Valence}} \\ 
\hline \hline
Modality& Model& $\mathcal{U}$ & $\mathcal{B}$ 
& $\mathcal{U}$ & $\mathcal{B}$  
\\ \hline
\multirow{3}{*}{Audio} 
& $M_I$ & 13.59 & \textbf{0.64} & 5.97 & \textbf{0.53}  \\ \cline{2-6}
& $M_S$ & 13.18 & \textbf{0.62} & 5.90 & \textbf{0.55}\\  \cline{2-6}
& $M_F$ & 13.77 & \textbf{0.64} & 5.87 & \textbf{0.57} \\ \hline\hline

\multirow{3}{*}{Fusion} 
& $M_I$ & 16.87 & \textbf{0.78} & 6.18 & \textbf{0.61}  \\ \cline{2-6}
& $M_S$ & 15.73 & \textbf{0.75} & 5.89 & \textbf{0.60}\\  \cline{2-6}
& $M_F$ & 16.00 & \textbf{0.76} & 6.46 & \textbf{0.63} \\ \hline\hline

\multicolumn{2}{c}{\textbf{SEWA}} & 
\multicolumn{2}{c}{\textbf{Arousal}} &
\multicolumn{2}{c}{\textbf{Valence}} \\ 
\hline \hline
Modality& Model& $\mathcal{U}$ & $\mathcal{B}$
& $\mathcal{U}$ & $\mathcal{B}$  
\\ \hline
\multirow{3}{*}{Visual} 
& $M_I$ & 1.98 & \textbf{1.08} & 2.35 & \textbf{0.65}  \\ \cline{2-6}
& $M_S$ & 2.03 & \textbf{0.91} & 2.10 & \textbf{0.87}\\  \cline{2-6}
& $M_F$ & 2.40 & \textbf{0.78} & 2.58 &\textbf{0.62} \\ \hline\hline
\multirow{3}{*}{Fusion} 
& $M_I$ & 2.12 & \textbf{1.98} & 1.77 & \textbf{1.68}  \\ \cline{2-6}
& $M_S$ & 1.77 & \textbf{1.24} & 1.69 & \textbf{0.98}\\  \cline{2-6}
& $M_F$ & 2.30 & \textbf{1.64} & 1.72 &\textbf{1.09} \\ \hline\hline

\end{tabular}
}
\label{table:kl}
\end{table}

\begin{table}[tbh!]
\centering
\caption{CCC performance for higher-order descriptors on RECOLA and SEWA. 
$B$: point-regressor baselines trained separately for each descriptor. 
$M$: proposed Beta-based approach deriving descriptors from $\mu,\sigma$.}
\centering
\resizebox{0.67\columnwidth}{!}{%
\begin{tabular}{l||@{ }c@{ }|@{ }c@{ }|@{ }c@{ }|@{ }c@{ }|@{ }c@{ }}
%

\multicolumn{2}{c}{\textbf{RECOLA}} & 
\multicolumn{2}{c}{\textbf{Arousal}} &
\multicolumn{2}{c}{\textbf{Valence}} \\ 
\hline \hline
Modality& Descriptor& $B$ & $M$ 
& $B$ & $M$  
\\ \hline
\multirow{5}{*}{Audio} 
& median & \underline{0.16} & \textbf{0.18} & \underline{0.50} & \textbf{0.52}  \\ \cline{2-6}
& q75 & 0.10 & \textbf{0.13} & \textbf{0.45} & \underline{0.44}\\  \cline{2-6}
& q25 & 0.12 & \textbf{0.18} & 0.50 & \textbf{0.53} \\ \cline{2-6}
& skew & \underline{-0.02} & \textbf{0.01} & -0.01 & \textbf{0.03}\\  \cline{2-6}
& kurt & \textbf{0.05} & -0.01 & \textbf{0.04} & \underline{0.02}\\  \hline\hline 

\multirow{5}{*}{Fusion} 
& median & \underline{0.30} & \textbf{0.31} & 0.44 & \textbf{0.48}  \\ \cline{2-6}
& q75 & \underline{0.23} & \textbf{0.25} & \underline{0.39} & \textbf{0.40}\\  \cline{2-6}
& q25 & \textbf{0.26} & \underline{0.24} & 0.48 & \textbf{0.51} \\ \cline{2-6}
& skew & \underline{0.02} & \textbf{0.02} & \underline{0.01} & \textbf{0.01}\\  \cline{2-6}
& kurt & \textbf{0.06} & 0.01 & \textbf{0.02} & \underline{0.01}\\  \hline\hline 

\multicolumn{2}{c}{\textbf{SEWA}} & 
\multicolumn{2}{c}{\textbf{Arousal}} &
\multicolumn{2}{c}{\textbf{Valence}} \\ 
\hline \hline
Modality& Model& $B$ & $M$ 
& $B$ & $M$  
\\ \hline

\multirow{5}{*}{Visual} 
& median & 0.24 & \textbf{0.30} & \underline{0.30} & \textbf{0.32}  \\ \cline{2-6}
& q75 & 0.35 & \textbf{0.41} & \textbf{0.53} & \underline{0.51}\\  \cline{2-6}
& q25 & \textbf{0.91} & 0.80 & \textbf{0.90} & 0.85 \\ \cline{2-6}
& skew & \underline{0.28} & \textbf{0.30} & \textbf{0.25} & \underline{0.23}\\  \cline{2-6}
& kurt & \underline{0.05} & \textbf{0.06} & \textbf{0.07} & \underline{0.04}\\  \hline\hline 

\multirow{5}{*}{Fusion} 
& median & \underline{0.27} & \textbf{0.27} & 0.30 & \textbf{0.34}  \\ \cline{2-6}
& q75 & \textbf{0.40} & \underline{0.38} & \textbf{0.54} & \underline{0.51}\\  \cline{2-6}
& q25 & \textbf{0.90} & 0.85 & \textbf{0.89} & 0.83 \\ \cline{2-6}
& skew & \textbf{0.48} & 0.30 & \textbf{0.33} & 0.28\\  \cline{2-6}
& kurt & \textbf{0.10} & \underline{0.08} & \textbf{0.12} & \underline{0.10}\\  \hline\hline

\end{tabular}
}
\label{table:descr}
\end{table}

\subsubsection{Prediction of Mean and Deviation}

Table~\ref{table:models} reports CCC results for predicting the mean ($\mu$) and standard deviation ($\sigma$) of annotation distributions across modalities on RECOLA and SEWA. Across both datasets, multimodal fusion consistently improves performance over unimodal inputs, highlighting the complementary nature of behavioural signals in affect modelling. Emotional expression is inherently multimodal, and combining cues from multiple sources appears particularly beneficial when modelling annotator consensus.

Dataset-specific modality patterns are also observed. In SEWA, visual features provide the strongest predictive signal. This behaviour is expected given that SEWA captures spontaneous interactions in naturalistic settings where facial behaviour plays a dominant role in conveying affective information. In contrast, RECOLA shows stronger contributions from audio features, reflecting the conversational nature of the collaborative task and the importance of vocal dynamics in expressing emotional activation during dialogue. These observations are consistent with previous studies that report modality-dependent performance differences across affect datasets.

Predicting the standard deviation $\sigma$ is notably more challenging than predicting the mean $\mu$, as $\sigma$ reflects annotator disagreement rather than a single perceptual signal. This suggests that modelling consensus variability requires capturing subtle cues related to emotional ambiguity or perceptual uncertainty. Nevertheless, the models are able to learn meaningful signals associated with annotation dispersion, indicating that behavioural features contain information not only about the central tendency of emotional perception but also about the degree of agreement among annotators.

Performance differences between the three architectures ($M_I$, $M_S$, and $M_F$) are generally small and often not statistically significant. This suggests that modelling annotator consensus through the proposed framework does not require complex architectures and can be achieved with relatively lightweight networks. The results therefore indicate that predicting the two distribution parameters $(\mu,\sigma)$ constitutes a stable and tractable learning objective across different modalities and datasets.

\subsubsection{Recovery of Higher-Order Distribution Descriptors}

Table~\ref{table:descr} evaluates the ability of the proposed framework to recover higher-order statistics of the annotation distribution. In this experiment, the Beta-based formulation ($M$) derives descriptors directly from the predicted $(\mu,\sigma)$ parameters and is compared with point-regressor baselines ($B$) trained independently for each descriptor.

Overall, the Beta-based approach performs on par with or better than the baseline models for most descriptors. Median and quartile estimates are recovered most reliably, indicating that the central structure of the annotation distribution can be reconstructed accurately from the predicted parameters. This behaviour is expected because these statistics depend primarily on the first moments of the distribution, which are captured directly by $\mu$ and $\sigma$.

Skewness proves moderately more challenging to estimate, while kurtosis is the most difficult descriptor to recover. Both statistics are highly sensitive to small variations in the underlying distribution and depend strongly on the precise relationship between $\alpha$ and $\beta$. As a result, small prediction errors in $\mu$ or $\sigma$ can propagate into larger deviations in these higher-order measures. Nevertheless, the proposed approach remains competitive with direct regression baselines despite requiring only two predicted parameters. This suggests that the Beta parameterisation provides a compact but expressive representation of annotator perception.

These results demonstrate that predicting only two parameters is sufficient to reconstruct a broad set of distributional properties. Instead of training separate models for each statistic, the proposed formulation provides a coherent representation of annotation distributions that captures both central tendency and higher-order characteristics. From a modelling perspective, this significantly reduces the number of learning targets while preserving rich information about annotator disagreement.

\subsubsection{Distributional Fidelity}

Finally, Table~\ref{table:kl} evaluates how closely the predicted distributions match the empirical annotation distributions using KL divergence. Ground-truth consensus distributions were approximated by fitting Beta distributions to the empirical annotation mean and variance, while predicted distributions were obtained by converting the predicted $(\hat{\mu},\hat{\sigma})$ values into Beta parameters. As a reference baseline, we use the uniform distribution $\mathcal{U}(0,1) \equiv \mathrm{Beta}(1,1)$. Across most settings, the predicted distributions are significantly closer to the empirical Beta distributions than to the uniform baseline. This indicates that the proposed models recover meaningful distributional structure rather than producing trivial or degenerate outputs. The improvement is particularly clear in RECOLA, where predicted distributions closely align with the empirical consensus. This suggests that the Beta assumption provides a reasonable approximation of annotation behaviour in controlled interaction settings.

For SEWA under multimodal fusion, the gap between predicted and empirical distributions is somewhat smaller. This observation is consistent with the slightly larger errors observed in the higher-order descriptor estimates. SEWA contains more diverse and spontaneous interactions across multiple cultural contexts, which likely introduces greater variability in emotional perception and annotation behaviour. Consequently, modelling the full structure of annotator disagreement in such in-the-wild data remains more challenging. Overall, the KL divergence analysis confirms that the proposed framework captures non-trivial distributional patterns in annotator responses. Rather than collapsing annotations into a single consensus value, the model learns distributions that reflect both agreement and ambiguity in emotional perception.

\subsubsection{Qualitative Assessment and Computational Efficiency}

\begin{figure}[htbp]
    \centering
    \begin{subfigure}[b]{0.48\textwidth}
        \centering
        \includegraphics[width=\textwidth]{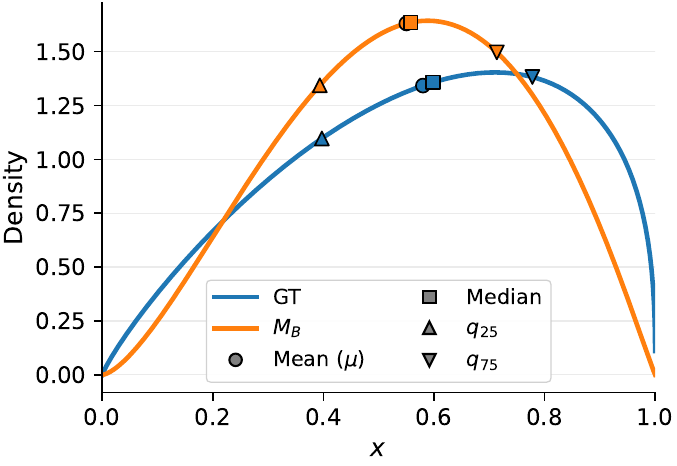}
        \caption{RECOLA: Arousal}
        \label{fig:rec_aro}
    \end{subfigure}
    \hfill
    \begin{subfigure}[b]{0.48\textwidth}
        \centering
        \includegraphics[width=\textwidth]{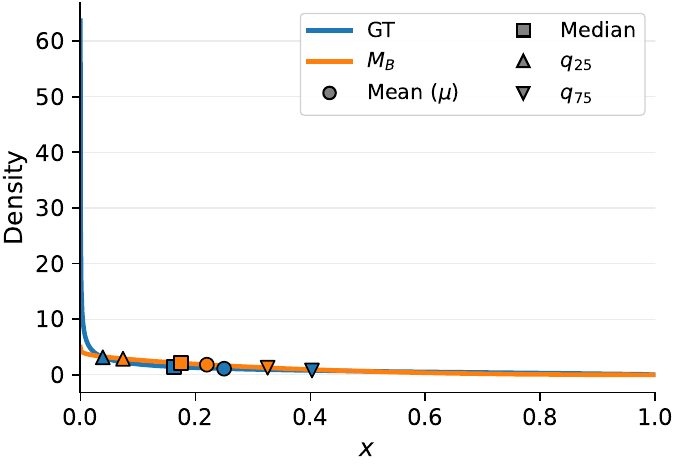}
        \caption{RECOLA: Valence}
        \label{fig:rec_val}
    \end{subfigure}

    \caption{Indicative true and predicted probability density functions for Beta-Based Regression ($M_B$).}
    \label{fig:distribution_shapes}
\end{figure}

To further assess the representational capacity of the Beta-based regression model ($M_B$), we qualitatively examine the predicted distribution shapes in comparison with the empirical expert consensus in the RECOLA dataset. Since the Beta distribution is defined on the bounded interval $[0,1]$, it naturally aligns with the normalised affect annotation space. Figure~\ref{fig:distribution_shapes} presents representative examples of predicted arousal and valence distributions, illustrating how $M_B$ approximates the underlying annotator responses. The predicted density functions demonstrate the flexibility of the Beta formulation in modelling different annotation patterns. In the valence dimension, the model successfully captures strongly skewed distributions that arise when annotators converge toward extreme affective interpretations. In contrast, arousal often exhibits broader and more symmetric distributions reflecting greater perceptual ambiguity. By reconstructing these patterns, the model provides richer information than a single scalar estimate, revealing not only the perceived emotional state but also the level of agreement among annotators.

Despite this representational flexibility, the current optimisation strategy introduces certain limitations. The model is trained by minimising a joint loss function composed of the Mean Squared Error (MSE) of the predicted mean ($\mu$) and standard deviation ($\sigma$). This formulation treats the two parameters as independent targets during optimisation, even though they are mathematically coupled in the Beta distribution through the relationship between mean and variance. As a result, the model may occasionally produce parameter combinations that are numerically valid but suboptimal from a probabilistic perspective, particularly near the boundaries of the affective space where the distribution becomes highly sensitive to small variations in $\sigma$. This issue is further reflected in the empirical results, where the Concordance Correlation Coefficient (CCC) for $\sigma$ tends to be lower than that for $\mu$. Because the loss function assigns equal weight to both objectives, the optimisation process may prioritise the more stable mean prediction while underfitting subtle variations in annotator disagreement. Future work could address this limitation by designing objective functions that explicitly enforce the structural coupling of Beta parameters or by optimising distributional likelihoods directly.

Beyond representational properties, the proposed framework remains computationally efficient. The underlying architecture is intentionally lightweight, consisting of a two-layer bottleneck network with a very small number of trainable parameters. Depending on the input modality configuration, the total parameter count ranges from approximately 1.8k to 19.3k. Importantly, extending the baseline regression model to the Beta-based formulation requires only a negligible increase in parameters at the output layer. Empirical benchmarking was performed on a standard workstation (11th Gen Intel Core i5 CPU, 16\,GB RAM, NVIDIA RTX 3060 GPU). This computational footprint indicates that the model can operate comfortably in real-time settings and could be deployed on modest hardware without requiring specialised acceleration. Consequently, the framework provides a practical balance between probabilistic modelling capability and computational efficiency.

\section{Discussion and Future Work}

While the proposed framework demonstrates that distribution-aware modelling can capture meaningful structure in emotion annotations, several limitations remain and open opportunities for future research. First, the current formulation assumes that annotation distributions can be approximated by the Beta family. Although the Beta distribution provides a flexible representation for bounded variables, it is inherently limited to unimodal or U-shaped distributions. In practice, annotator responses may exhibit more complex structures, including multimodal patterns that arise when different annotator groups interpret the same behaviour differently \cite{makantasis2022invariant}. Extending the framework to mixture-based or non-parametric distributions could therefore provide a more expressive representation of annotator disagreement.

Second, the estimation of Beta parameters relies on moment matching from the empirical mean and variance of annotations. While this approach is simple and computationally efficient, it can be sensitive to noise in cases where annotation variance is extremely small or when annotations lie near the boundaries of the label range. Future work could explore alternative parameter estimation techniques, such as likelihood-based optimisation or direct prediction of Beta parameters with appropriate regularisation. Another limitation concerns the prediction of the annotation variance $\sigma^2$, which represents the degree of disagreement between annotators. Unlike the mean, which reflects a central perceptual tendency, disagreement arises from multiple sources, including stimulus ambiguity, perceptual differences between annotators, and annotation delays. Modelling these factors explicitly—for example through annotator-specific representations or uncertainty-aware architectures—could improve the prediction of distributional dispersion.

The current study also collapses annotator responses into aggregate statistics without modelling annotator identity or reliability. In practice, annotators often exhibit systematic biases or temporal response differences. Incorporating annotator-aware modelling strategies, such as reliability weighting, annotator embeddings, or hierarchical probabilistic models, could provide deeper insights into the sources of disagreement. Another limitation lies in the experimental scope. The proposed framework is evaluated on RECOLA and SEWA, two well-established benchmarks for continuous emotion prediction. While these datasets offer rich multimodal recordings, they contain relatively limited annotator pools and structured interaction scenarios. Evaluating distribution-aware modelling on larger and more diverse datasets, such as Aff-Wild2 or MuSe, could provide further evidence of the generality of the approach.

Finally, the present study employs lightweight feedforward neural networks and independent temporal windows. While this design emphasises methodological simplicity, emotion perception is inherently dynamic. Future work could integrate temporal modelling architectures such as transformer-based networks to capture longer-range affect dynamics and annotation lag effects. Overall, these limitations highlight several promising directions for advancing distribution-aware affect modelling. By combining richer probabilistic representations with improved temporal modelling and annotator-aware learning, future research may further enhance the ability of machine learning systems to capture the inherently subjective nature of human emotional perception. Importantly, the proposed framework is not restricted to affective computing; its formulation is general and can be readily applied to other domains involving subjective, noisy, and multi-annotator signals, such as player and viewer engagement modelling \cite{pinitas2024across,pinitas2023predicting}, where capturing variability and disagreement is equally critical.

\section{Conclusions}

This work introduces a Beta-based framework for modelling annotator consensus in continuous affect prediction. By predicting the mean and standard deviation of annotation signals, the approach reconstructs full probabilistic distributions that capture variability, asymmetry, and higher-order properties such as skewness and quantiles. Experiments on the RECOLA and SEWA datasets show that these two parameters are sufficient to recover rich distributional information while maintaining competitive predictive performance. The results highlight the importance of modelling subjective annotations as distributions rather than single-valued targets and demonstrate that lightweight models can effectively learn such representations. Although evaluated on affective computing tasks, the framework is general and applicable to other domains involving bounded and ambiguous human annotations.

\section*{Acknowledgements}

This work is co-funded by the HORIZON-CL4-2023-HUMAN-01-CNECT, Grant Agreement number: 101135990 - Human-centric Digital Twin Approaches to Trustworthy AI and Robotics for Improved Working Conditions (AI4Work).